\RequirePackage{fix-cm}
\documentclass[twocolumn]{svjour3}          
\smartqed  
\usepackage[utf8]{inputenc} 
\usepackage[T1]{fontenc}    
\usepackage{hyperref}       
\usepackage{url}            
\usepackage{booktabs}       
\usepackage{amsfonts}       
\usepackage{nicefrac}       
\usepackage{microtype}      
\usepackage{graphicx}
\usepackage{epstopdf}
\usepackage{subfigure}
\usepackage{enumitem}

\usepackage{xcolor}

\usepackage{amsmath}
\usepackage{multirow}
\usepackage[T1]{fontenc}
\usepackage{lmodern}
\begin{document}

\title{An Effective Dynamic Spatio-temporal Framework \\ with Multi-Source Information for Traffic Prediction
}


\author{Jichen Wang \and
        Weiguo Zhu \and
        Yongqi Sun$^{*}$ \and
        Chunzi Tian
}


\institute{Jichen Wang \and Weiguo Zhu \and Yongqi Sun \and Chunzi Tian \at
              School of Computer and Information Technology, Beijing Jiaotong University, Beijing 100044, P. R. China
              \\\email{yqsun@bjtu.edu.cn}
}

\date{}

\maketitle

\begin{abstract}
Traffic prediction is necessary not only for management departments to dispatch vehicles but also for drivers to avoid congested roads. Many traffic forecasting methods based on deep learning have been proposed in recent years, and their main aim is to solve the problem of spatial dependencies and temporal dynamics. In this paper, we propose a useful dynamic model to predict the urban traffic volume by combining fully bidirectional LSTM, the more complex attention mechanism, and the external features, including weather conditions and events. First, we adopt the bidirectional LSTM to obtain temporal dependencies of traffic volume dynamically in each layer, which is different from the hybrid methods combining bidirectional and unidirectional ones; second, we use a more elaborate attention mechanism to learn short-term and long-term periodic temporal dependencies; and finally, we collect the weather conditions and events as the external features to further improve the prediction precision. The experimental results show that the proposed model improves the prediction precision by approximately 3-7 percent on the NYC-Taxi and NYC-Bike datasets compared to the most recently developed method, being a useful tool for the urban traffic prediction.
\end{abstract}

\section{Introduction}
\label{intro}
In recent years, due to the surge in people's car holdings and limited road carrying capacity, the contradiction between road supply and demand has become prominent increasingly. The traffic volume is one of the main parameters reflecting the running state of the road. If the traffic volume on the road can be monitored and predicted in time and accurately, the vehicle can be guided in advance expertly. This will not only improve the operational capability and operational efficiency of the road network but also be of considerable significance to traffic managers, operators, and participants.

Over the past three decades, there has been a dramatic increase in traffic forecasting research. Different methods~\cite{Cui:2018traffic} have been proposed for traffic prediction, and those methods can be divided into three categories, traditional time series, traditional machine learning, and deep learning techniques. The first category includes autoregressive integrated moving average (ARIMA)~\cite{Williams:2003br}, season autoregressive integrated moving average(SARIMA)~\cite{Szeto:2009bi}, and Kalman filtering~\cite{Xie:2007je,Guo:2010fb} with its variants, and they have already been used in many traffic prediction applications. However, these classical statistical methods require that the input data must meet certain premise assumptions. Therefore it is not sufficient for them to process complex and non-linear traffic data, and cannot model the real circumstance meticulously~\cite{Yao:2019tz}. Researchers add extra information, such as human activity, weather data, and holiday plan to those methods and make a better decision~\cite{Wu:2016vp,Tong:2017vy,Zhang:2017tq,Ye:2019cp}.

The second category includes traditional machine learning approaches, such as KNN~\cite{van2012short,Luo:2019hm} and SVM~\cite{Jeong:ww,Sun:2015wj}. Traditional machine learning approaches and traditional time series methods are similar in data processing. It is more useful if the types and features of traffic data are given, but it is challenging to set the parameters in advance~\cite{Guo2019AttentionBS}. Moreover, these simple machine learning methods can model more complex traffic data, but they do not take into account the temporal and spatial dependencies of traffic data simultaneously. Therefore, deep learning techniques are more useful and accessible for traffic prediction.

Recently, deep learning has achieved the best precision in many challenging tasks, such as the game of go~\cite{Silver:2016tu}, speech recognition, object detection, and many other domains~\cite{LeCun:2015wc}. Affected by the great success of deep learning, more and more researchers are trying to use deep learning techniques to traffic prediction. For example, Lv et al. use stacked autoencoders to learn generic traffic flow features~\cite{Lv:2015dl}. Cui et al. transfer the city-wide traffic to a heatmap image to catch the non-linear spatial dependency by using the convolutional neural network (CNN)~\cite{Cui:2018deep,Yao:2018wy,Zhang:2018uj}. At the same time, some researchers make use of the recurrent neural network (RNN)~\cite{Yao:2019tz,Cui:2018traffic,Cui:2018deep} to catch non-linear temporal dependency. A ST-ResNet model designed base on the residual convolution unit to predict traffic flow~\cite{Zhang:2017tq}. Besides, graph convolutional network(GCN) and graph neural network(GNN) as the supplement to CNN have been applied to the traffic flow prediction together with CNN~\cite{Yu:2018gn,Zhao:2018vv,Guo2019AttentionBS}. A Spatial-Temporal Dynamic Network(STDN) is proposed to learn the similarity between locations dynamically for taxi demand prediction~\cite{Yao:2018wy}, and the STDN is put forward to solve dynamic problems~\cite{Yao:2019tz}.

The traffic volume forecast results are somewhat dependent on long-term periodicity. The prediction of traffic volume is not only affected by the data dependency of the previous item, but also the situation of traffic prediction should affect the previous prediction result to some extent. In addition to the above factors, weather conditions have a tremendous impact on the forecast of traffic volume, and the impact level of weather conditions on different traveling choices is also inconsistent. For example, heavy rain has a more critical impact on bicycles than on buses or taxis.

In this paper, we propose a novel framework based on a more elaborate attention mechanism and bidirectional LSTM combined with weather conditions and other external features. We use the proposed framework to conduct the experiments on the NYC-Taxi and NYC-Bike datasets. The experimental results show that the prediction accuracy of our framework is better than the existing baselines.
The main contributions of this paper are as follows:
\begin{itemize}
	\item We utilize a bidirectional LSTM based on two-way feature dependency to extract omnidirectional features of traffic volume.
	\item We propose a more elaborate multi-scale attention mechanism that combines long-term and short-term periodic dynamic temporal dependency acquisition.
	\item We add the external features such as weather, holidays, weekdays, and weekends into our model to improve the prediction accuracy of traffic volume.
\end{itemize}

\section{Related Work}
\label{relatedwork}
Traffic volume is operating the traffic data as a consecutive fluid~\cite{Greenberg:1959hg}. Traffic volume includes not only the operations of the vehicles but also the activities of bicycles and pedestrians. The traffic volume forms a time series dynamical with time and is embedded in the continuous space, and thus traffic volume forecasting is a typical spatio-temporal data mining problem. In this section, we focus on the work related to traffic volume forecasting.

In the traditional time series prediction methods, ARIMA \cite{Williams:2003br} and SARIMA \cite{Szeto:2009bi} have been applied to some linear data prediction modules. However, the accuracy of these methods for non-linear traffic data cannot be guaranteed. Although the KNN~\cite{van2012short} and SVM \cite{Jeong:ww} methods are capable of handling more complex data, it is necessary to know the data types and features and set the corresponding parameters in advance. Affected by the great success of deep learning, more and more researchers use deep learning techniques to predict traffic data. A wide variety of networks are designed and applied by researchers in traffic volume prediction. Although several methods, such as MLFNN\cite{Dougherty:1997vt}, RNN \cite{Yasdi:1999wt,Zhang:2000dx}, and WNN \cite{BotoGiralda:2010jd}, can model the non-linear traffic data to some extent, their network structures only consider the partial temporal dependency of traffic data.

To solve this problem, CNN has been used to predict traffic volume in recent years~\cite{Yu:2017tv,Yao:2018wy,Zhang:2018uj}. Shi et al. proposed the LSTM extension, the ConvLSTM, which is better than FC-LSTM and suitable for the spatio-temporal data~\cite{Shi:2015uua}. Three residual neural networks are composed dynamically for forecasting the flow of crowds~\cite{Zhang:2017tq}.
Similarly, a highly flexible and extendible end-to-end framework, DeepSD is proposed for modeling the car-hailing service supply-demand~\cite{Wang:2017bt}. DeepSD combined with the deep residual network and embedding method is used for learning the patterns by spatial-temporal attributes. They regard the entire city as a map and divide the entire city equally into $n*m$ areas or according to traffic checkpoints, such as bus stops, traffic lights, and highway toll stations. These algorithms take into account both spatial-temporal dependencies but do not consider the dynamic relationship of spatial-temporal dependence of traffic data.

Yao et al. designed a multi-view spatial-temporal network (DMVST-Net)~\cite{Yao:2018wy}, which combines spatial attributes, temporal attributes, and semantic attributes. The local CNN and graph embedding methods are used in DMVST-Net to get more views of data. Moreover, they further designed the deep learning network and proposed the framework called Spatial-Temporal Dynamic Network (STDN)~\cite{Yao:2019tz}, which can extract the similarity between different regions dynamically. However, their method still makes insufficient consideration of dynamic temporal dependency. For example, the traffic volume in the morning peaks can affect the traffic conditions in the evening peaks, and the traffic conditions in different seasons are different. Besides, most methods do not take into account the impact of external features such as weather, holidays, weekdays, weekends, and seasons on traffic data.

In summary, we propose a novel model that is different from the literatures. Our framework is based on a new multi-scale attention mechanism, forward and backward dependency information on traffic data and the external features to improve the prediction accuracy.

\begin{figure*}[ht]
\centering
\includegraphics[width=0.8\textwidth]{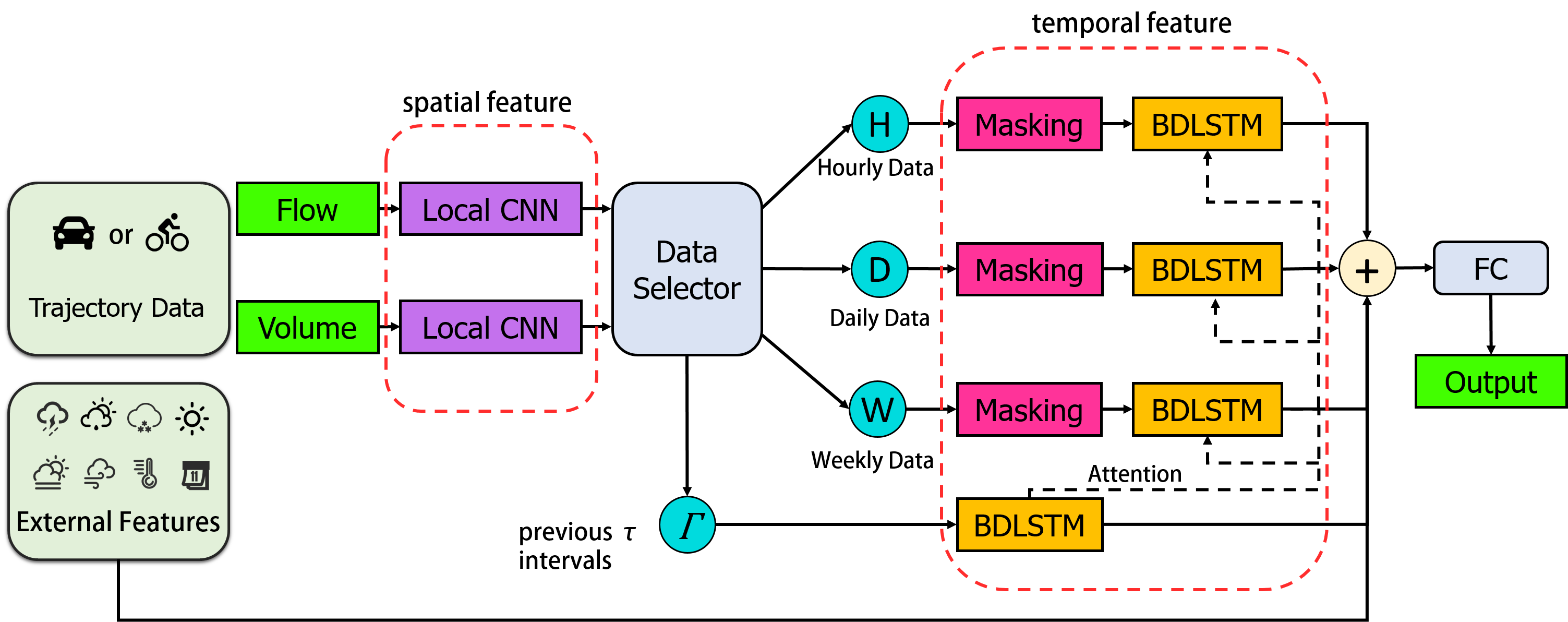}
\caption{The framework of the network. Input data include traffic flow, traffic volume, and external features. The traffic features are extracted by local CNN and BDLSTM with an attention mechanism. The external features are combined with the obtained traffic data features, and the final output is obtained through a fully connected layer.}
\label{netStruc}
\end{figure*}

\section{Methods}
In this section, we will focus on the proposed network framework. The overall network framework is shown in Figure \ref{netStruc}. The traffic volume represents the number of vehicles arriving or departing a region within a fixed time interval. The traffic flow represents the number of vehicles entering into other regions within a fixed time interval. Traffic flow reflects the association between different regions.

\subsection{Local CNN} We use local CNN to obtain local spatial dependency. The main idea is to weaken the reliance between the weaker correlation region and the prediction region, and enhance the dependence between the strongly correlated region and the prediction region \cite{Yao:2018wy,Yao:2019tz}. Experiments have shown that regions with weaker correlations with predicted regions regularly reduce the accuracy of traffic prediction.

For each time interval $t$, we handle the target region $i$ and its surrounding neighbors as an image $\mathcal{F}_{i,t} \in \mathbb{R}^{S*S*2}$ with two channels. The first channel represents traffic conditions at the start of the time interval, and the other channel represents traffic conditions at the end of the time interval. The target region is the center point of the image. Then, local CNN takes $\mathcal{F}_{i,t}$ as input $\mathcal{F}_{i,t}^{0}$, and the expression of each convolution layer $k$ is:
\begin{equation}
	\mathcal{F}_{i, t}^{k}=f\left(W_{t}^{k}*\mathcal{F}_{i, t}^{k-1} + b_{t}^{k}\right)
\end{equation}
\noindent
, where $*$ represents the convolution operation, $W_t^{k}$ and $b_t^{k}$ are learned parameters. After a total of $k$ convolutions, the features $\mathcal{F}_{i, t}$ of the target region $i$ are obtained and transmitted to the fully connected layer.

\subsection{Masking Mechanism}
\begin{figure}[ht]
\centering
\includegraphics[width=0.95\columnwidth]{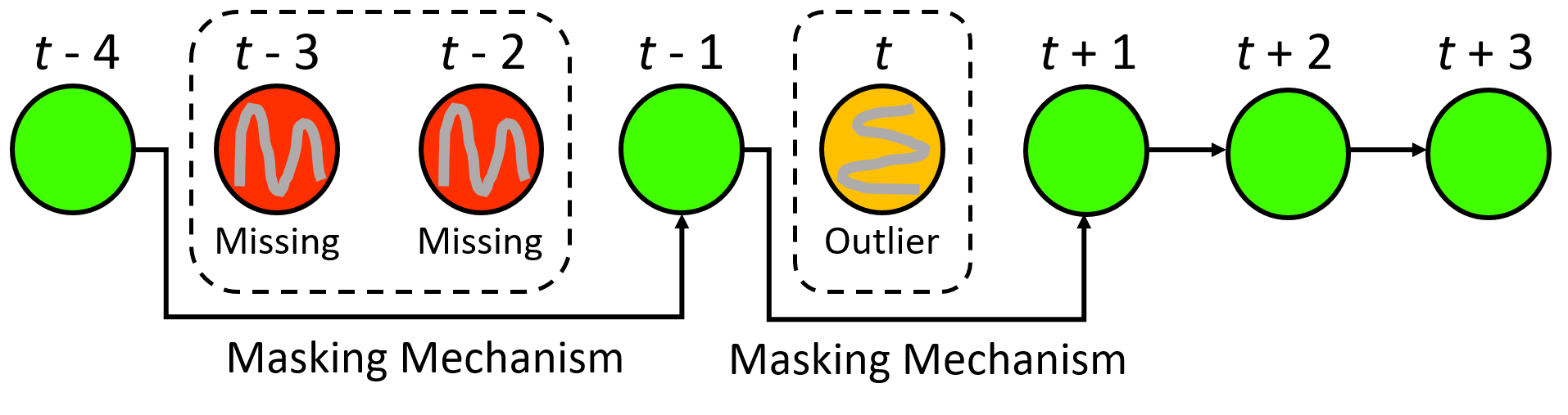}
\caption{Masking mechanism for traffic volume with missing values and outliers.}
\label{mask}
\end{figure}

There may be missing values or outliers which are less than zero or greater than 0.5 in the traffic data, and the bidirectional LSTM (BDLSTM) model cannot train with these values. If we assign them to zero or the fixed values, the prediction accuracy of the framework would be significantly reduced. Similar to \cite{Cui:2018deep}, we use a masking mechanism to reduce the impact of these values on model training. The operation of the masking mechanism is shown in Figure \ref{mask}. Initially, we assume that all data is reasonable. Then, they are fed into local CNN, and precessed by the masking operation on the convolution output. For example, if the data at time $t$ is a missing value or outlier, the BDLSTM training process of step $t$ is skipped, and the result of step $t-1$ is fed into step $t+1$. If the data at time $t+1$ is still a missing value or outlier, the processing is skipped again until the value is reasonable. In this way, the missing values or outliers need not be set to zero or the fixed values, and thus the BDLSTM training process is not affected by them.

\begin{figure}[ht]
\centering
\includegraphics[width=0.95\columnwidth]{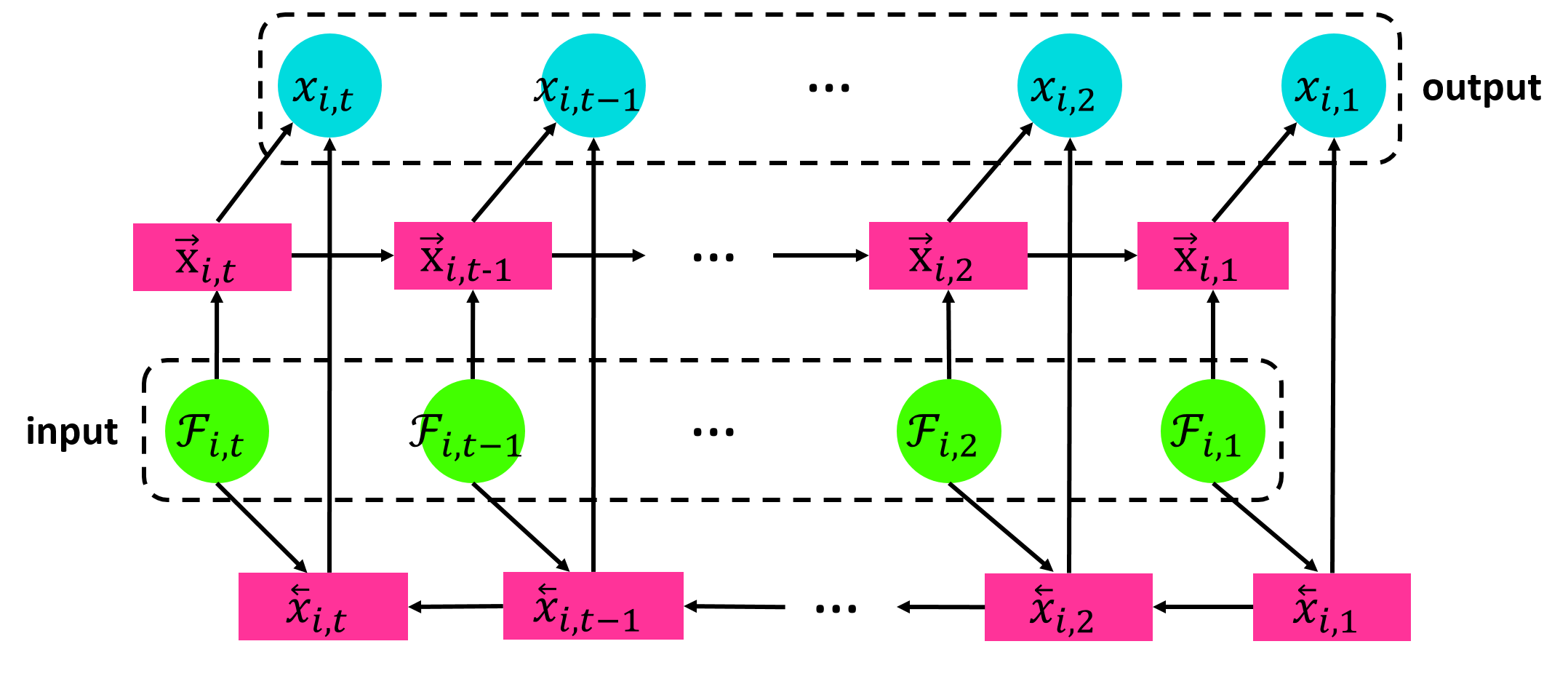} 
\caption{Detailed architecture of BDLSTM.}
\label{BDLSTM}
\end{figure}

\subsection{BDLSTM}
We use a BDLSTM network \cite{Cui:2018deep} to obtain the temporal dependency of traffic data, as shown in Figure \ref{BDLSTM}. The main reasons for using BDLSTM are as follows:
\begin{itemize}
	\item Generally, the input data of the LSTM is arranged in the temporal order. However, LSTM only utilizes the positive dependency of the data, which may cause useful features to be filtered out or not pass the chain-like gated structure effectively.
	\item Besides, analyzing the temporal periodicity of traffic data from a forward and backward perspective can help us find the recurring traffic patterns and improve prediction precision.
	\item In fact, traffic data does not increase or decrease suddenly. We can utilize BDLSTM to smooth the data so that the predicted results are closer to the ground truth.
\end{itemize}

The BDLSTM mentioned in this paper contains a forward LSTM and a backward LSTM. We use $\stackrel{\rightarrow}{x}_{i, t}$ and $\stackrel{\leftarrow}{x}_{i, t}$ to represent the inputs iteratively calculated in a positive sequence and a reversed sequence for the target region $i$ at time $t$, respectively. The output $x_{i, t}$ is defined as follow:
\begin{equation}
	x_{i, t}=BDLSTM\left(\mathcal{F}_{i, t}, \stackrel{\rightarrow}{x}_{i, t}, \stackrel{\leftarrow}{x}_{i, t}\right)
\end{equation}
, where $\mathcal{F}_{i, t}$ represent the features of the target region $i$ at time $t$.

\subsection{Attention Mechanism}
Typically, scholars adopt an attention mechanism to capture the temporal shifting in the long-term traffic volume of the daily or weekly periodicity~\cite{Yao:2019tz,Guo2019AttentionBS}. To improve the prediction accuracy, we add more information to the attention mechanism, which includes three kinds of time intervals, hourly (e.g., $\tau$ intervals), daily, and weekly periodicity. For the hourly level, the traffic data between time $t-\tau$ and time $t$ is used to predict the traffic volume of time $t$. For the daily (resp. weekly) level, the traffic data between time $t-\tau$ and time $t+\tau$ of the previous $d$ days (resp. $w$ weeks) is used to predict the traffic volume of time $t$.

\begin{figure*}[t]
\centering
\includegraphics[width=0.9\textwidth]{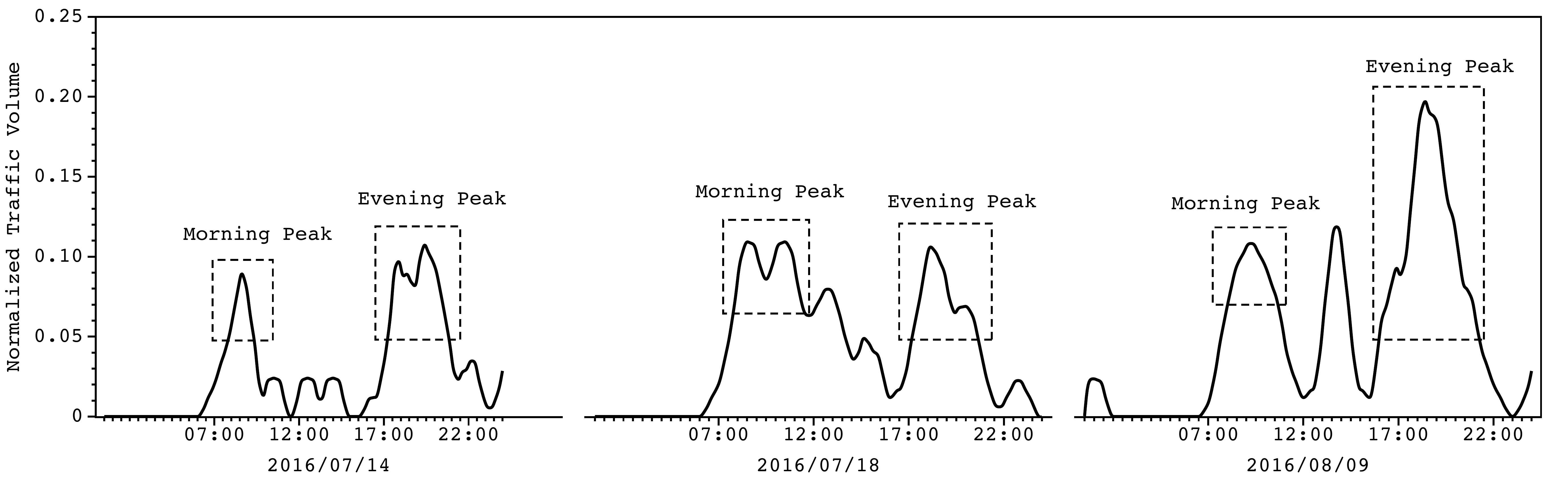}
\caption{Traffic volume of three days selected randomly.}
\label{peak}
\end{figure*}

Moreover, through a large amount of data analysis and comparative experiments, we found that there are strong relationships between traffic conditions at the morning peak and the ones at the evening peak each day, as shown in Figure \ref{peak}.
So, it is useful for predicting the traffic volume between 20:00 and 20:30 to consider the traffic volume around 9:00.
Therefore, we regard the traffic volume before $h$ hours as an essential reliance on specific periods to improve the prediction accuracy of the model. We use BDLSTM to learn the correlation of these different levels, which are defined as follows:
\begin{equation}
	\begin{aligned}
		x_{i, t}^{h, \tau} &=\operatorname{BDLSTM}\left(\mathcal{F}_{i, t}^{h,\tau}, \stackrel{\rightarrow}{x}_{i, t}^{h,\tau-1}, \stackrel{\leftarrow}{x}_{i, t}^{h,\tau-1}\right) \\
		x_{i, t}^{d, \tau} &=\operatorname{BDLSTM}\left(\mathcal{F}_{i, t}^{d,\tau}, \stackrel{\rightarrow}{x}_{i, t}^{d,\tau-1}, \stackrel{\leftarrow}{x}_{i, t}^{d,\tau-1}\right) \\
		x_{i, t}^{w, \tau} &=\operatorname{BDLSTM}\left(\mathcal{F}_{i, t}^{w,\tau}, \stackrel{\rightarrow}{x}_{i, t}^{w,\tau-1},  \stackrel{\leftarrow}{x}_{i, t}^{w,\tau-1}\right)
	\end{aligned}
\end{equation}
, where $i$ represents the target region, $t$ represents the time to be predicted,
$x_{i, t}^{h, \tau}$ represents the output of the ${t}-{h}\pm{\tau/2}$ hours period ($\tau$ is a interval of half an hour),
$x_{i, t}^{d, \tau}$ (resp. $x_{i, t}^{w, \tau}$) represents the output of the ${t}\pm{\tau/2}$ period before $d$ days (resp. $w$ weeks).
$\mathcal{F}_{i, t}^{h,\tau}$ represents the input of the ${t}-{h}\pm{\tau/2}$ hours period.
$\mathcal{F}_{i, t}^{d,\tau}$ (resp. $\mathcal{F}_{i, t}^{w,\tau}$) represents the input of the ${t}\pm{\tau/2}$ hours period before $d$ days (resp. $w$ weeks).
We use an attention mechanism to capture the dynamic temporal dependency, and obtain the weighted representations of the previous $h$ hours, $d$ days and $w$ weeks as follows:
\begin{equation}
	\begin{aligned}
		x_{i, t}^{h} &=\sum_{\tau \in \varGamma} \alpha_{i, t}^{h, \tau} x_{i, t}^{h, \tau} \\
 		x_{i, t}^{d} &=\sum_{\tau \in \varGamma} \alpha_{i, t}^{d, \tau} x_{i, t}^{d, \tau} \\
  		x_{i, t}^{w} &=\sum_{\tau \in \varGamma} \alpha_{i, t}^{w, \tau} x_{i, t}^{w, \tau}
  	\end{aligned}
\end{equation}
, where $x_{i, t}^{h}$, $x_{i, t}^{d}$, and $x_{i, t}^{w}$ represent the weighted sum of the output features of the previous $h$ hours, $d$ days, and $w$ weeks corresponding to the region $i$ at time $t$, respectively. The weights $\alpha_{i, t}^{h, \tau}$, $\alpha_{i, t}^{d, \tau}$ and $\alpha_{i, t}^{w, \tau}$ are used to measure the importance of the time intervals ${h}$, ${d}$ and ${w}$, respectively, and defined as follows:
\begin{equation}
\begin{aligned}
	\alpha_{i, t}^{h, \tau}&=\frac{\exp (\operatorname{score}(x_{i, t}^{h, \tau}, x_{i, t}))}{\sum_{h' \in \mathbb{H}} \exp (\operatorname{score}(x_{i, t}^{h', \tau}, x_{i, t}))} \\
 	\alpha_{i, t}^{d, \tau}&=\frac{\exp (\operatorname{score}(x_{i, t}^{d, \tau}, x_{i, t}))}{\sum_{d' \in \mathbb{D}} \exp (\operatorname{score}(x_{i, t}^{d', \tau}, x_{i, t}))} \\
  	\alpha_{i, t}^{w, \tau}&=\frac{\exp (\operatorname{score}(x_{i, t}^{w, \tau}, x_{i, t}))}{\sum_{w' \in \mathbb{W}} \exp (\operatorname{score}(x_{i, t}^{w', \tau}, x_{i, t}))}
  \end{aligned}
\end{equation}
The $score$ function is defined as:
\begin{small}
\begin{equation}
\begin{aligned}
	\operatorname{score}\left(x_{i, t}^{h, \tau}, \!x_{i, t}\right)\!&=\!V_{1}^{T}\!\tanh\!\left(W_{\gamma_{1}}\!x_{i, t}^{h, \tau}\!+\!W_{\beta1} x_{i, t}\!+\!b_{\beta1}\right) \\
	\operatorname{score}\left(x_{i, t}^{d, \tau}, \!x_{i, t}\right)\!&=\!V_{2}^{T}\!\tanh\!\left(W_{\gamma_{2}}\!x_{i, t}^{d, \tau}\!+\!W_{\beta2} x_{i, t}\!+\!b_{\beta2}\right) \\
 	\operatorname{score}\left(x_{i, t}^{w, \tau}, \!x_{i, t}\right)\!&=\!V_{3}^{T}\!\tanh\!\left(W_{\gamma_{3}}\!x_{i, t}^{w, \tau}\!+\!W_{\beta3} x_{i, t}\!+\!b_{\beta3}\right)
\end{aligned}
\end{equation}
\end{small}
, where $W_{\gamma_{1}}$, $W_{\beta_{1}}$, $W_{\gamma_{2}}$, $W_{\beta_{2}}$, $W_{\gamma_{3}}$, $W_{\beta_{3}}$, $b_{\beta_{1}}$, $b_{\beta_{2}}$, $b_{\beta_{3}}$, $V_{1}$, $V_{2}$, and $V_{3}$ are learned parameters, and $V^{T}$ is a transposition of $V$.
For the previous $h$ hours, $d$ days, and $w$ weeks, we obtain the values of ${x}_{i, t}^{h}$, ${x}_{i, t}^{d}$, and ${x}_{i, t}^{w}$, respectively. Then, we use these representations as the input of the following BDLSTM to preserve the periodic information,
\begin{equation}
\begin{aligned}
	\hat{x}_{i, t}^{h} &=\operatorname{BDLSTM}\left(x_{i, t}^{h}, \stackrel{\rightarrow}{\hat{x}}_{i, t}^{h-1}, \stackrel{\leftarrow}{\hat{x}}_{i, t}^{h-1}\right) \\
 	\hat{x}_{i, t}^{d} &=\operatorname{BDLSTM}\left(x_{i, t}^{d}, \stackrel{\rightarrow}{\hat{x}}_{i, t}^{d-1}, \stackrel{\leftarrow}{\hat{x}}_{i, t}^{d-1}\right) \\
  	\hat{x}_{i, t}^{w} &=\operatorname{BDLSTM}\left(x_{i, t}^{w}, \stackrel{\rightarrow}{\hat{x}}_{i, t}^{w-1}, \stackrel{\leftarrow}{\hat{x}}_{i, t}^{w-1}\right)
\end{aligned}
\end{equation}
The final outputs of $\hat{x}_{i, t}^{h}$, $\hat{x}_{i, t}^{d}$, and $\hat{x}_{i, t}^{w}$ represent the dynamic temporal dependence.
\subsection{Integration with External Features}
We first capture the weather conditions of New York City from the website\footnote{https://darksky.net} and extract the other external features, including holidays, weekdays, and weekends. This multiple information is pre-processed and used as the external features of the target region $i$ at time $t$, denoted as $external_{i,t}$. Then, we integrate the short-term output $x_{i, t}$ with the three long-term output $\hat{x}_{i, t}^{h}$, $\hat{x}_{i, t}^{d}$ and $\hat{x}_{i, t}^{w}$ together, denoted as $x_{i, t}^{c}$, where we adopt the shorter intervals in the short-term function to capture the detail information. Finally, $x_{i, t}^{c}$ is combined with the external features $external_{i,t}$, and the prediction results $f_{i, t+1}^{s}$ and $f_{i, t+1}^{e}$ are obtained through a fully connected layer, which are defined as follows:
\begin{small}
\begin{equation}
\left[f_{i, t+1}^{s},\!f_{i, t+1}^{e}\right]\!=\!\tanh\!\left(W_{full}\!\left[x_{i, t}^{c},\!external_{i, t}\right]\!+\!b_{full}\right)
\end{equation}
\end{small}
, where $W_{full}$ and $b_{full}$ are the parameters learned. $f_{i, t+1}^{s}$ and $f_{i, t+1}^{t}$ indicate the prediction results of the target area $i$ at the start and the end of at time $t+1$ respectively.

\section{Experiment}
\subsection{Experimental Data}
We use two datasets, NYC-Taxi and NYC-Bike, to verify the effectiveness of our prediction model, their information is shown in Table \ref{tab:datasets}. For the NYC-Taxi dataset, there are 22,349,490 taxi trip records from 01/01/2015 to 03/01/2015. For the NYC-Bike dataset, there are 2,605,648 trip records from 07/01/2016 to 08/29/2016. For the two datasets, we choose the last twenty days data as the testing data which plays the same role of the validation data and other data as training data. The external features we collected include weather conditions, holidays, weekends, and weekdays. The details of the external features are shown in Table \ref{tab:external}.
\begin{table}[ht]
  \centering
  \caption{The datasets of the NYC-Bike and NYC-Taxi.}
    \begin{tabular}{lll}
    \hline
    Dataset & NYC-Taxi & NYC-Bike \\
    \hline
    \multirow{2}{*}{Time span} & 01/01/2015- & 07/01/2016- \\
     & 03/01/2015 & 08/29/2016 \\
    Time interval & \multicolumn{2}{c}{30 minutes} \\
    Gird map size & \multicolumn{2}{c}{(10, 20)} \\
    \hline
    \end{tabular}%
  \label{tab:datasets}%
\end{table}%
\begin{table}[ht]
	\centering
	\caption{The external features of the NYC-Bike and NYC-Taxi.}
	\begin{tabular}{lll}
		\hline
		Dataset & NYC-Taxi & NYC-Bike \\
		\hline
		Temperature/$^{\circ}$F                                &  [1, 53] & [58, 95] \\
		Wind speed/mph                              &  [0, 17] &  [0, 8] \\
		Humidity/\%                                   & [30, 89] & [31, 95] \\
		UV index                                           & [0,4] & [0,10] \\
		Precip probability/\%                             & [0, 100] & [0, 100] \\
		Armospheric pressure/mb                         &  [991, 1044] & [1005, 1027] \\
		Visibility/mi                                     & [0, 11] & [3, 10] \\
		holidays                     &  3 & 1\\
		weekend                           &  1$\sim$7 & 1$\sim$7\\
		\hline
	\end{tabular}%
	\label{tab:external}%
\end{table}%
\subsection{Preprocessing}
In the beginning, we normalize the training data to [0, 1] by using Max-Min normalization on the training set. In the evaluation, we re-scale the predicted values back to the standard values and compare them with the ground truth. For the external features, we first use the de-unit process to remove the information of units and then normalize them to [0-1] by way of Max-Min. Besides, we use one-hot coding to transform holiday, weekday, and weekend conditions into binary vectors.

\subsection{Parameters}
The python library Keras is used to construct the BDLSTM in our method. We set up a 3-layer convolution to extract the spatial dependency of the traffic volume, the size of each layer of convolution kernel is 3*3 with 64 filters, the dimension of hidden representation of BDLSTM is 128. For the length of the short-time traffic data, we set $\tau=3$, namely the previous 1.5 hours. For the length of the long-time traffic data, we set the corresponding peak $h=11$ (i.e., from previous 9.5 to 12.5 hours), the number of days $d = 3$ (i.e., the previous three days), the number of weeks $w=1$ (i.e., previous one week). We choose 2/3 of the data as the training set and the remaining data as the test set. The batch size is 128, and the learning rate is $10e-5$. The optimizer used in our experiment is Adagrad. We finetune the CNN network after 20 epochs. We set the batch size to be 80 and train for up to 150 epochs with early stopping if the loss score had not improved over the last 6 epochs.

\subsection{Evaluation Methods}
We use the metrics of Rooted Mean Square Error (RMSE) and Mean Average Percentage Error (MAPE) to evaluate our proposed model, which are defined as follows:

\begin{equation}
\operatorname{RMSE}=\sqrt{\frac{1}{n} \sum_{i=1}^{n} \left(\hat{f}_{i, t+1}-f_{i, t+1}\right)^{2}}
\end{equation}

\begin{equation}
\operatorname{MAPE}=\frac{1}{n} \sum_{i=1}^{n} \frac{\left|\hat{f}_{i, t+1}-f_{i, t+1}\right|}{f_{i, t+1}}
\end{equation}
, where $n$ is the total number of samples, $f_{i,t+1}$ and $\hat{f}_{i,t+1}$ represent the ground truth and the prediction result, respectively.

\subsection{Baselines}
We compare our model with the following conventional methods that employ a neural network to predict traffic volume.
\begin{itemize}
\item {ConvLSTM~\cite{Shi:2015uua}:} It uses convolution to obtain the spatial dependency of the data and then adopts LSTM to obtain temporal dependencies.
\item {DeepSD~\cite{Wang:2017bt}:} An end-to-end deep learning framework that automatically learns the spatial-temporal features.
\item{ST-ResNet~\cite{Zhang:2017tq}:} ST-ResNet uses a convolution-based residual network to model the spatial-temporal dependency of any two regions in a city.
\item{DMVST-Net~\cite{Yao:2018wy}:} A unified multi-view model that jointly considers the spatial, temporal, and semantic relations.
\item{STDN~\cite{Yao:2019tz}:} STDN uses the flow gating mechanism to obtain the spatial dynamic dependency of traffic data, and obtain the dynamic dependency on temporal through the attention mechanism.
\end{itemize}
\section{Results and Discussion}
In this section, we will analyze the experimental results on the datasets of NYC-Bike and NYC-Taxi.
\subsection{Results on NYC-Bike}
The experimental results on the NYC-Bike dataset are shown in Table \ref{tab:bike}. It includes a comparison of the proposed model with the five baselines mentioned above including the latest method STDN, where there are three cases in the ablation experiment of our model.
We can know by comparison that the different factors we added show varying degrees of influence on the overall experimental results. Overall, compared to the latest method STDN, our model reduces the RMSE and MAPE values by 5.99\% and 3.43\% at the start of the period, respectively. At the end of the period, it reduces the RMSE and MAPE values by 6.01\% and 5.03\%, respectively.
\begin{table*}[htbp]
\setlength{\abovecaptionskip}{0pt}
\centering
\caption{Experimental results on the NYC-Bike dataset. The results of ConvLSTM, DeepSD, ST-ResNet, DMVST-Net, and STDN are refered from~\cite{Yao:2019tz}.}\smallskip
\begin{tabular}{lllll}
\hline
model & RMSE-START & MAPE-START & RMSE-END & MAPE-END\\
\hline
ConvLSTM   & $10.40$           & $25.10\%$           & $9.22$           & $23.20\%$ \\
DeepSD     & $9.69$            & $23.62\%$           & $9.08$           & $22.36\%$ \\
ST-ResNet  & $9.80$            & $25.06\%$           & $8.85$           & $22.98\%$ \\
DMVST-Net  & $9.14$            & $22.20\%$           & $8.50$           & $21.56\%$ \\
STDN       & $8.85$            & $21.84\%$           & $8.15$           & $20.87\%$ \\
\hline
LSTM+attention & $8.49$	           & $21.19\%$            & $8.07$           & $20.26\%$ \\
LSTM+external  & $8.45$            & $21.21\%$            & $8.11$           & $20.87\%$ \\
BDLSTM	       & $8.57$	           & $21.26\%$            & $7.76$           & $20.31\%$ \\
Our model  & $\textbf{8.32}$      & $\textbf{21.09\%}$  & $\textbf{7.66}$           & $\textbf{19.82\%}$ \\
\hline
\end{tabular}
\label{tab:bike}
\end{table*}

\begin{table*}[htbp]
    \setlength{\abovecaptionskip}{0pt}
	\centering
	\caption{Experimental results for different periods on the NYC-Bike dataset.}
	\begin{tabular}{llllll}
	    \hline
	    Predict Period  & Model  & RMSE-START  & RMSE-END  & MAPE-START  & MAPE-END\\
		\hline
		\multirow{5}{*}{Weekend}
	    & STDN                & $8.51$           & $8.10$           & $21.69\%$           & $20.78\%$ \\
		& LSTM+attention      & $8.36(\boldsymbol{-1.75\%})$  & $7.96(-1.65\%)$  & $21.27\%(-1.93\%)$  & $20.15\%(\boldsymbol{-3.05\%})$ \\
		& LSTM+external       & $8.37(-1.66\%)$  & $7.85(-3.95\%)$  & $21.23\%(\boldsymbol{-2.13\%})$  & $20.76\%(-0.08\%)$ \\
		& BDLSTM              & $8.39(-1.34\%)$  & $7.73(\boldsymbol{-4.53\%})$  & $21.29\%(-1.83\%)$  & $20.33\%(-2.19\%)$ \\
		& Our model           & $\boldsymbol{8.33(-2.11\%)}$  & $\boldsymbol{7.70(-4.92\%)}$  & $\boldsymbol{21.18\%(-2.38\%)}$  & $\boldsymbol{20.07\%(-3.44\%)}$ \\
		\hline
		\multirow{5}{*}{Weekdays}
		& STDN                & $9.08$           & $8.19$           & $21.94\%$           & $20.93\%$ \\
		& LSTM+attention      & $8.58(-5.44\%)$  & $8.13(-0.74\%)$   & $21.14\%(\boldsymbol{-3.64\%})$  & $20.33\%(-2.86\%)$ \\
		& LSTM+external       & $8.51(\boldsymbol{-6.29\%})$  & $8.18(-0.11\%)$   & $21.20\%(-3.36\%)$  & $20.93\%(-0.00\%)$ \\
		& BDLSTM              & $8.69(-4.25\%)$  & $7.77(\boldsymbol{-5.03\%})$  & $21.23\%(-3.24\%)$  & $20.33\%(\boldsymbol{-2.90\%})$ \\
		& Our model           & $\boldsymbol{8.28(-8.80\%)}$  & $\boldsymbol{7.63(-6.75\%)}$ & $\boldsymbol{20.92\%(-4.63\%)}$  & $\boldsymbol{19.77\%(-5.54\%)}$ \\
		\hline
		\multirow{5}{*}{Off-peak period}
		& STDN                & $9.04$            & $8.28$            & $22.08\%$           & $21.36\%$ \\
		& LSTM+attention      & $8.86(-2.00\%)$   & $8.13(-1.74\%)$   & $21.46\%(-2.83\%)$  & $20.41\%(\boldsymbol{-4.47\%})$ \\
		& LSTM+external       & $8.53(\boldsymbol{-5.70\%})$   & $8.15(-1.49\%)$   & $21.41\%(\boldsymbol{-3.05\%})$  & $21.02\%(-1.58\%)$ \\
		& BDLSTM              & $8.84(-2.23\%)$   & $7.93(\boldsymbol{-4.25\%})$   & $21.45\%(-2.88\%)$  & $20.70\%(-3.08\%)$ \\
		& Our model           & $\boldsymbol{8.42(-6.94\%)}$   & $\boldsymbol{7.67(-7.31\%)}$   & $\boldsymbol{21.38\%(-3.18\%)}$  & $\boldsymbol{20.36\%(-4.71\%)}$ \\
		\hline
		\multirow{5}{*}{Peak period}
		& STDN                & $8.38$            & $7.96$            & $21.25\%$           & $20.29\%$ \\
		& LSTM+attention      & $8.37(-0.10\%)$   & $7.91(-0.63\%)$   & $20.55\%(\boldsymbol{-3.30\%})$  & $19.89\%(\boldsymbol{-1.97\%})$ \\
		& LSTM+external       & $8.26(-1.43\%)$   & $7.94(-0.19\%)$   & $20.60\%(-3.08\%)$  & $20.12\%(-0.87\%)$ \\
		& BDLSTM              & $8.16(\boldsymbol{-2.67\%})$   & $7.63(\boldsymbol{-4.07\%})$   & $20.63\%(-2.92\%)$  & $19.90\%(-1.93\%)$ \\
		& Our model           & $\boldsymbol{8.09(-3.49\%)}$   & $\boldsymbol{7.55(-5.17\%)}$   & $\boldsymbol{20.54\%(-3.37\%)}$  & $\boldsymbol{19.71\%(-2.86\%)}$ \\
		\hline
		\multirow{5}{*}{All period}
		& STDN                & $8.85$            & $8.15$            & $21.84\%$           & $20.87\%$ \\
		& LSTM+attention      & $8.49(-4.02\%)$   & $8.07(-1.00\%)$   & $21.19\%(\boldsymbol{-2.96\%})$  & $20.26\%(\boldsymbol{-2.93\%})$ \\
		& LSTM+external       & $8.45(\boldsymbol{-4.52\%})$   & $8.11(-0.46\%)$   & $21.21\%(-2.87\%)$  & $20.87\%(-0.00\%)$ \\
		& BDLSTM              & $8.57(-3.13\%)$   & $7.76(\boldsymbol{-4.83\%})$   & $21.26\%(-2.68\%)$  & $20.31\%(-2.66\%)$ \\
		& Our model           & $\boldsymbol{8.32(-5.99\%)}$   & $\boldsymbol{7.66(-6.01\%)}$   & $\boldsymbol{21.09\%(-3.43\%)}$  & $\boldsymbol{19.82\%(-5.03\%)}$ \\
		\hline
	\end{tabular}%
	\label{tab:bike-dif-period-tbl}%
\end{table*}%
In order to further analyze the advantages of our model, we conduct a comparative experiment with STDN at special time intervals, e.g., peak hours versus off-peak hours, weekends versus weekdays. The experimental results are shown in Table \ref{tab:bike-dif-period-tbl}, where the values in parentheses are the relative error increments.
We can observe that our framework shows the best prediction performance on the weekends and weekdays, in the off-peak periods and peak periods separately. In addition, each factor we added has a different influence on prediction accuracy. From the data in Table \ref{tab:bike-dif-period-tbl}, we can see that the multi-scale attention mechanism has the most significant impact on forecasting traffic conditions on the weekends. BDLSTM has a greater impact on the weekdays, and external features have a greater impact on the off-peak periods. BDLSTM and multi-scale attention mechanisms have a more significant effect on the peak periods.

\begin{figure}[ht]
\centering
\includegraphics[width=0.95\columnwidth]{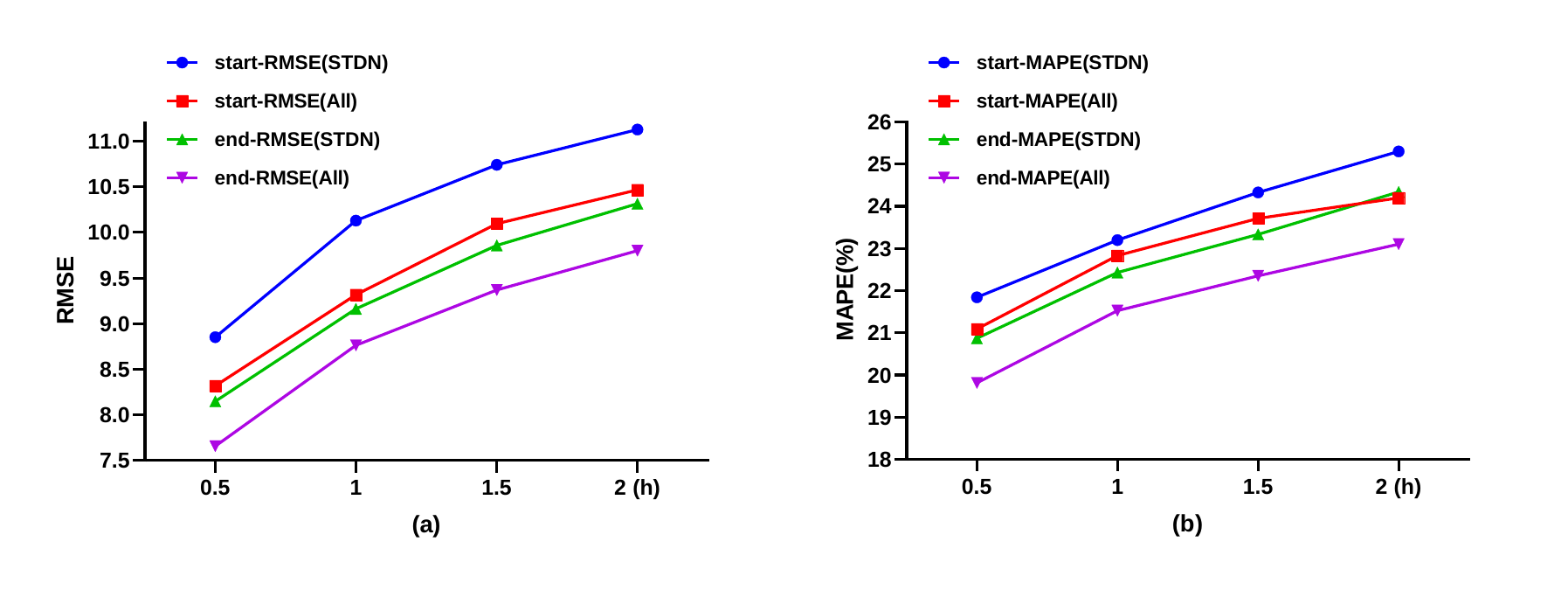} 
\caption{The experimental results of our model for different prediction intervals on the NYC-Bike dataset. Figures $(a)$ and $(b)$ indicate the RMSE and MAPE from 0.5 to 2 hours at the start and end, respectively.}
\label{bike-result-dif-interval}
\end{figure}

\begin{table*}[!ht]
\setlength{\abovecaptionskip}{0pt}
\centering
\caption{Experimental results on the NYC-Taxi dataset. The results of ConvLSTM, DeepSD, ST-ResNet, DMVST-Net, and STDN are refered from~\cite{Yao:2019tz}.}\smallskip
\begin{tabular}{lllll}
\hline
model & RMSE-START & MAPE-START &
RMSE-END & MAPE-END\\
\hline
ConvLSTM      & $28.13$ & $20.50\%$ & $23.67$ & $20.70\%$  \\
DeepSD        & $26.35$ & $18.12\%$ & $21.95$ & $18.15\%$  \\
ST-ResNet     & $26.23$ & $21.13\%$ & $21.63$ & $21.09\%$  \\
DMVST-Net     & $25.74$ & $17.38\%$ & $20.51$ & $17.14\%$  \\
STDN          & $24.10$ & $16.30\%$ & $19.05$ & $16.25\%$  \\
\hline
LSTM+attention & $23.69$     & $15.93\%$     & $18.78$      & $16.15\%$ \\
LSTM+external  & $22.49$     & $15.88\%$     & $17.81$      & $15.69\%$ \\
BDLSTM         & $23.57$     & $15.79\%$     & $17.93$      & $15.77\%$ \\
Our model  & $\textbf{22.37}$ & $\textbf{15.45\%}$   & $\textbf{17.78}$  & $\textbf{15.52\%}$ \\
\hline
\end{tabular}
\label{tab:taxi}
\end{table*}

\begin{table*}[!ht]
	\centering
	\caption{Experimental results for different periods on the NYC-Taxi dataset.}
	\begin{tabular}{llllll}
	    \hline
	    Predict Period  & Model  & RMSE-START  & RMSE-END  & MAPE-START  & MAPE-END\\
		\hline
		\multirow{5}{*}{Weekend}
	    & STDN                & $25.08$           & $20.13$           & $16.36\%$           & $16.31\%$ \\
		& LSTM+attention      & $24.17(-3.65\%)$  & $20.08(-0.24\%)$  & $16.31\%(-0.31\%)$  & $16.31\%(-0.00\%)$ \\
		& LSTM+external       & $23.99(-4.33\%)$  & $18.22(-9.50\%)$  & $16.22\%(-0.85\%)$  & $15.85\%(-2.82\%)$ \\
		& BDLSTM              & $23.82(\boldsymbol{-5.02\%})$  & $18.02(\boldsymbol{-10.45\%})$ & $15.94\%(\boldsymbol{-2.59\%})$  & $15.80\%(\boldsymbol{-3.16\%})$ \\
		& Our model           & $\boldsymbol{23.77(-5.22\%)}$ & $\boldsymbol{17.92(-10.98\%)}$ & $\boldsymbol{15.92\%(-2.70\%)}$ & $\boldsymbol{15.67\%(-3.95\%)}$ \\
		\hline
		\multirow{5}{*}{Weekdays}
		& STDN                & $23.45$           & $18.74$           & $16.26\%$           & $16.23\%$ \\
		& LSTM+attention      & $23.17(-1.20\%)$  & $18.61(-0.71\%)$  & $15.90\%(-2.22\%)$  & $16.13\%(-0.59\%)$ \\
		& LSTM+external       & $22.33(\boldsymbol{-4.78\%})$  & $17.79(\boldsymbol{-5.09\%})$  & $15.85\%(-2.55\%)$  & $15.66\%(\boldsymbol{-3.49\%})$ \\
		& BDLSTM              & $23.40(-0.19\%)$  & $17.87(-4.65\%)$  & $15.77\%(\boldsymbol{-3.01\%})$  & $15.76\%(-2.88\%)$ \\
		& Our model           & $\boldsymbol{22.24(-5.16\%)}$  & $\boldsymbol{17.69(-5.60\%)}$  & $\boldsymbol{15.44\%(-5.04\%)}$  & $\boldsymbol{15.45\%(-4.78\%)}$ \\
		\hline
		\multirow{5}{*}{Off-peak period}
		& STDN                & $23.75$           & $19.16$           & $16.31\%$           & $16.26\%$ \\
		& LSTM+attention      & $23.66(-0.37\%)$  & $18.92(-1.27\%)$  & $15.98\%(-2.02\%)$  & $16.23\%(-0.15\%)$ \\
		& LSTM+external       & $23.22(\boldsymbol{-2.23\%})$  & $17.81(\boldsymbol{-7.05\%})$  & $15.96\%(-2.15\%)$  & $15.87\%(-2.41\%)$ \\
		& BDLSTM              & $23.41(-1.46\%)$  & $17.99(-6.10\%)$  & $15.82\%(\boldsymbol{-3.00\%})$  & $15.86\%(\boldsymbol{-2.44\%})$ \\
		& Our model           & $\boldsymbol{23.27(-2.02\%)}$  & $\boldsymbol{17.80(-7.10\%)}$  & $\boldsymbol{15.74\%(-3.54\%)}$  & $\boldsymbol{15.77\%(-2.99\%)}$ \\
		\hline
		\multirow{5}{*}{Peak period}
		& STDN                & $24.94$           & $18.78$           & $16.27\%$           & $16.23\%$ \\
		& LSTM+attention      & $23.76(-4.74\%)$  & $18.67(-0.56\%)$  & $15.83\%(-2.68\%)$  & $16.14\%(-0.54\%)$ \\
		& LSTM+external       & $23.63(\boldsymbol{-5.25\%})$  & $17.82(-5.08\%)$  & $15.59\%(\boldsymbol{-4.17\%})$  & $15.55\%(\boldsymbol{-4.23\%})$ \\
		& BDLSTM              & $23.97(-3.89\%)$  & $17.78(\boldsymbol{-5.31\%})$  & $15.70\%(-3.51\%)$  & $15.64\%(-3.65\%)$ \\
		& Our model           & $\boldsymbol{21.99(-11.83\%)}$ & $\boldsymbol{17.73(-5.58\%)}$ & $\boldsymbol{15.38\%(-5.47\%)}$  & $\boldsymbol{15.36\%(-5.38\%)}$ \\
		\hline
		\multirow{5}{*}{All period}
		& STDN                & $24.10$           & $19.05$           & $16.30\%$           & $16.25\%$ \\
		& LSTM+attention      & $23.69(-1.70\%)$  & $18.78(-1.40\%)$  & $15.93\%(-2.25\%)$  & $16.15\%(-0.61\%)$ \\
		& LSTM+external       & $22.49(\boldsymbol{-6.67\%})$  & $17.81(\boldsymbol{-6.49\%})$  & $15.88\%(-2.57\%)$  & $15.69\%(\boldsymbol{-3.46\%})$ \\
		& BDLSTM              & $23.57(-2.20\%)$  & $17.93(-5.87\%)$  & $15.79\%(\boldsymbol{-3.08\%})$  & $15.77\%(-2.95\%)$ \\
		& Our model           & $\boldsymbol{22.37(-7.18\%)}$  & $\boldsymbol{17.78(-6.67\%)}$  & $\boldsymbol{15.45\%(-5.21\%)}$  & $\boldsymbol{15.52\%(-4.49\%)}$ \\
		\hline
	\end{tabular}%
	\label{tab:taxi-dif-period-tbl}%
\end{table*}%

In order to test the performance of our framework at different time intervals, we select four prediction scenarios with prediction periods of 0.5h, 1h, 1.5h, and 2h, respectively. Figure \ref{bike-result-dif-interval} shows the RMSE and MAPE of STDN and our model in different prediction intervals. As the prediction interval increases, the experimental error gradually increases from 0.5 to 2 hours. It is worth noting that as the prediction interval of our framework increases, the rate of error rise (i.e., the slope of the red and purple polylines) is significantly lower than that of STDN. The result shows that our framework is useful not only in the short-term prediction interval but also in the case of long-term traffic condition prediction.

\subsection{Results on NYC-Taxi}

The experimental results on the NYC-Taxi dataset are shown in Table \ref{tab:taxi}. Similarly, it includes a comparison of the proposed model with the five baselines mentioned above including the latest method STDN, where there are three cases in the ablation experiment of our model. From the experimental results in Table \ref{tab:taxi}, we can see that the three factors have played a decisive role in improving prediction accuracy.
In particular, compared with the latest method STDN, our model reduces the RMSE and MAPE values by 7.18\% and 5.21\% at the start of the period, respectively. At the end of the period, it reduces the RMSE and MAPE values by 6.67\% and 4.49\%, respectively.

\begin{figure}[htb]
\centering
\includegraphics[width=0.95\columnwidth]{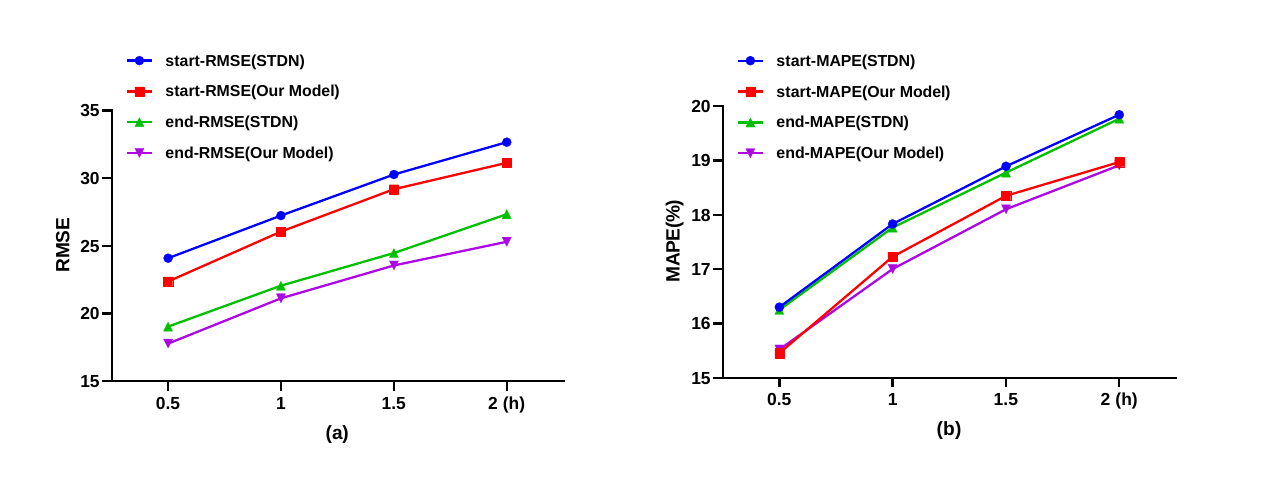} 
\caption{The experimental results of our model for different prediction intervals on the NYC-Taxi dataset. Figures $(a)$ and $(b)$ indicate the RMSE and MAPE from 0.5 to 2 hours at the start and end, respectively.}
\label{taxi-result-dif-interval}
\end{figure}

Similarly, in order to further analyze the advantages of our model, we conduct a comparative experiment with STDN at special time intervals, e.g., peak hours versus off-peak hours, weekends versus weekdays. The experimental results are shown in Table \ref{tab:taxi-dif-period-tbl}. From it, we can see that for the NYC-Taxi dataset, BDLSTM has a higher degree of reduction in the prediction error rate for the weekends and off-peak periods. External features have a higher degree of influence on weekdays, off-peak periods, and peak periods. In addition, when making predictions for all periods, external features can greatly reduce prediction errors.

Similarly, Figure \ref{taxi-result-dif-interval} shows the RMSE and MAPE of STDN and our model in different prediction intervals. As the prediction interval increases, the experimental error from 0.5 to 2 hours gradually increases. It is worth noting that as the prediction interval of our network framework increases, the rate of error rise (i.e., the slope of the red and purple polylines) is significantly lower than that of STDN.

\subsection{Effects of Different Components}
\begin{figure}[ht]
\centering
\includegraphics[width=0.95\columnwidth]{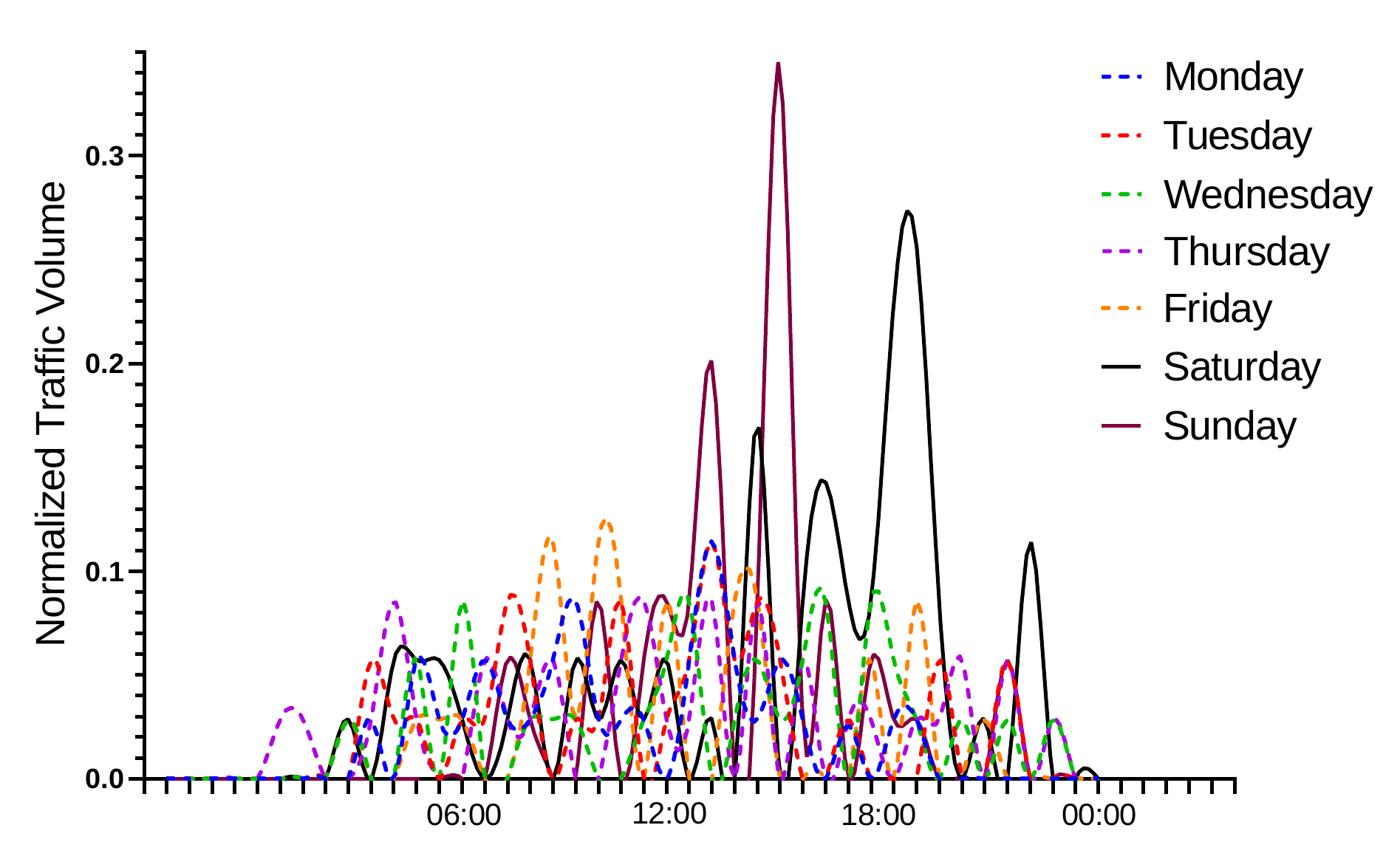} 
\caption{Normalized traffic volume of one random week. The dotted lines represent the data of weekday, and the solid lines represent the data of weekend.}
\label{external-everyday}
\end{figure}

\begin{figure}[ht]
\centering
\includegraphics[width=0.95\columnwidth]{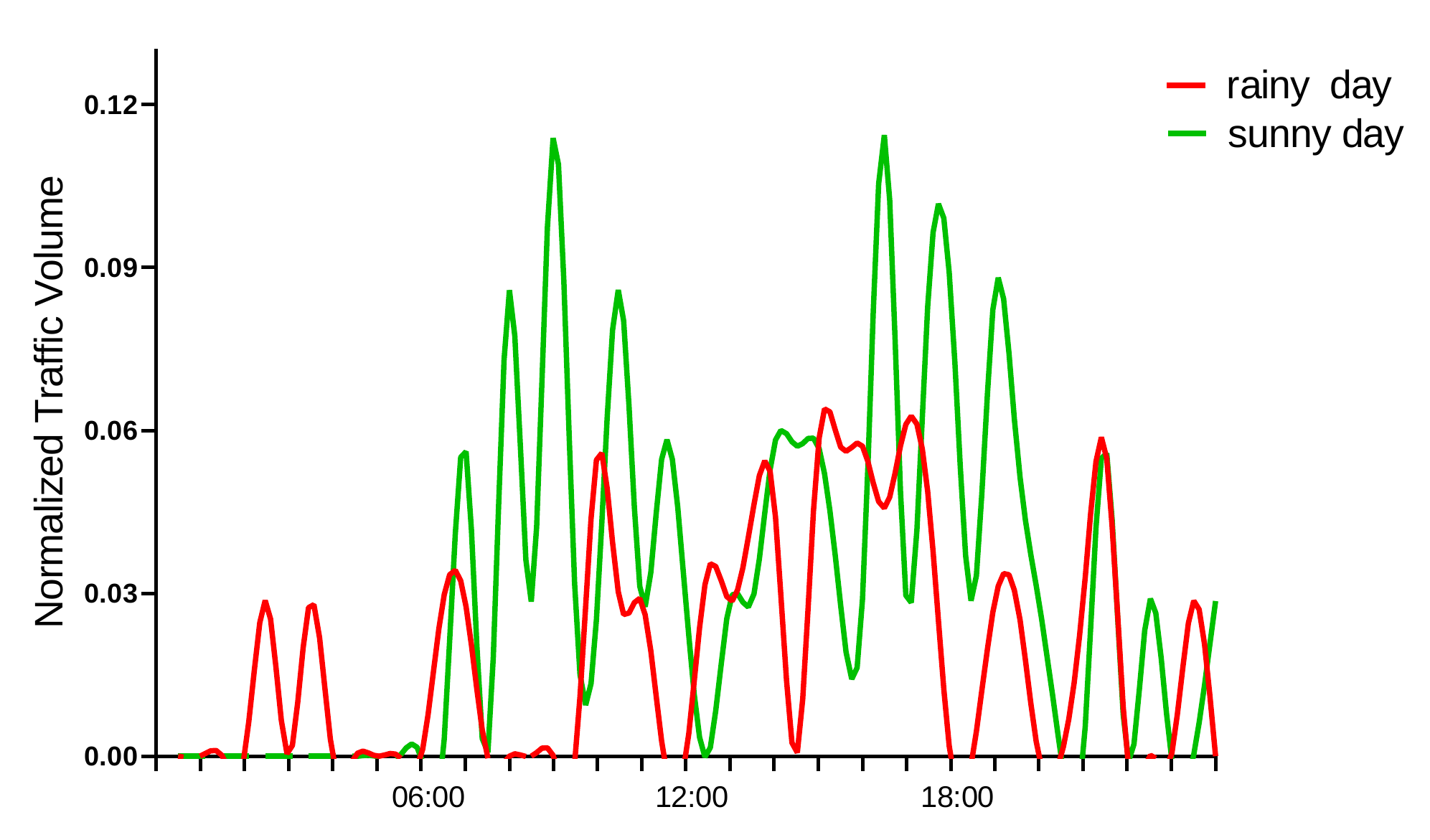} 
\caption{Normalized traffic volume of sunny and rainy days.}
\label{external-rain-sunny}
\end{figure}

 \textbf{External Features:} From Tables  \ref{tab:bike}, \ref{tab:bike-dif-period-tbl}, \ref{tab:taxi}, and \ref{tab:taxi-dif-period-tbl}, we can conclude that the external characteristics (weather conditions, holidays, etc.) have a notable influence on improving the predicting accuracy in the short-term traffic condition. For the factor of holidays, we randomly select the traffic conditions for a week, as shown in Figure \ref{external-everyday}. We can see that the traffic volume on the weekends fluctuates significantly, while the traffic volume on weekdays maintains a potential regularity. In addition, for the factor of weather conditions (rainy and sunny days), we randomly select the traffic volume for two weekdays, as shown in Figure \ref{external-rain-sunny}. We can see that the trends of traffic volume on rainy and sunny days are significantly different.

\begin{figure}[ht]
\centering
\includegraphics[width=0.95\columnwidth]{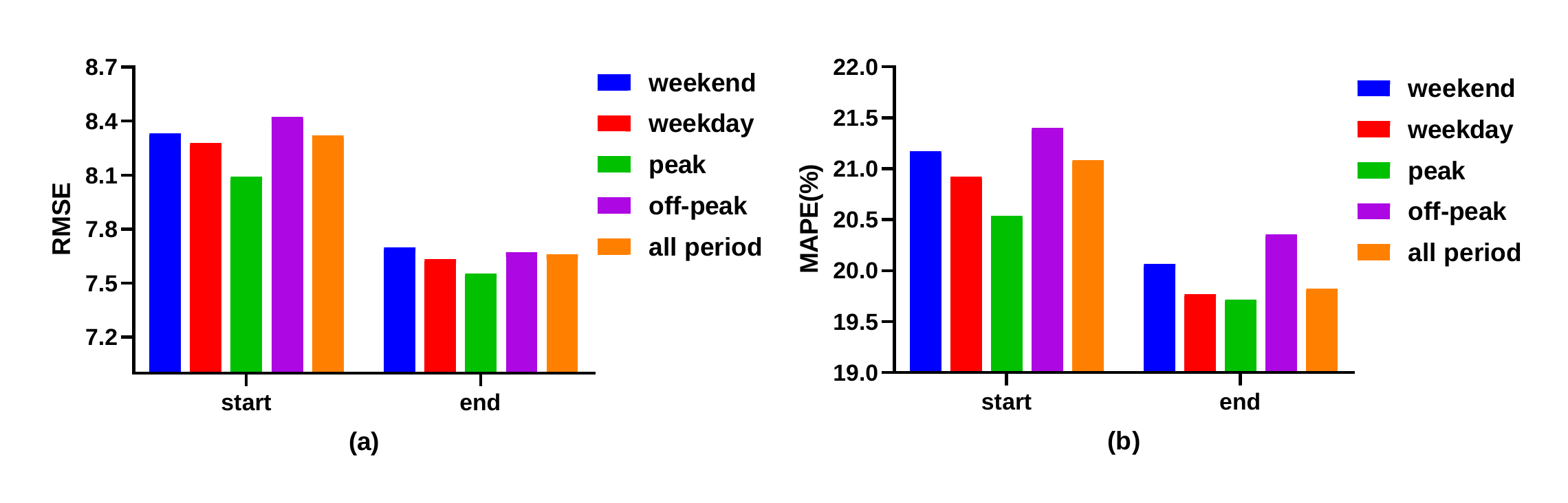} 
\caption{The experimental results of our integrated framework for different prediction periods on the NYC-Bike dataset.}
\label{bike-result-all-dif-period}
\end{figure}

\begin{figure}[ht]
\centering
\includegraphics[width=0.95\columnwidth]{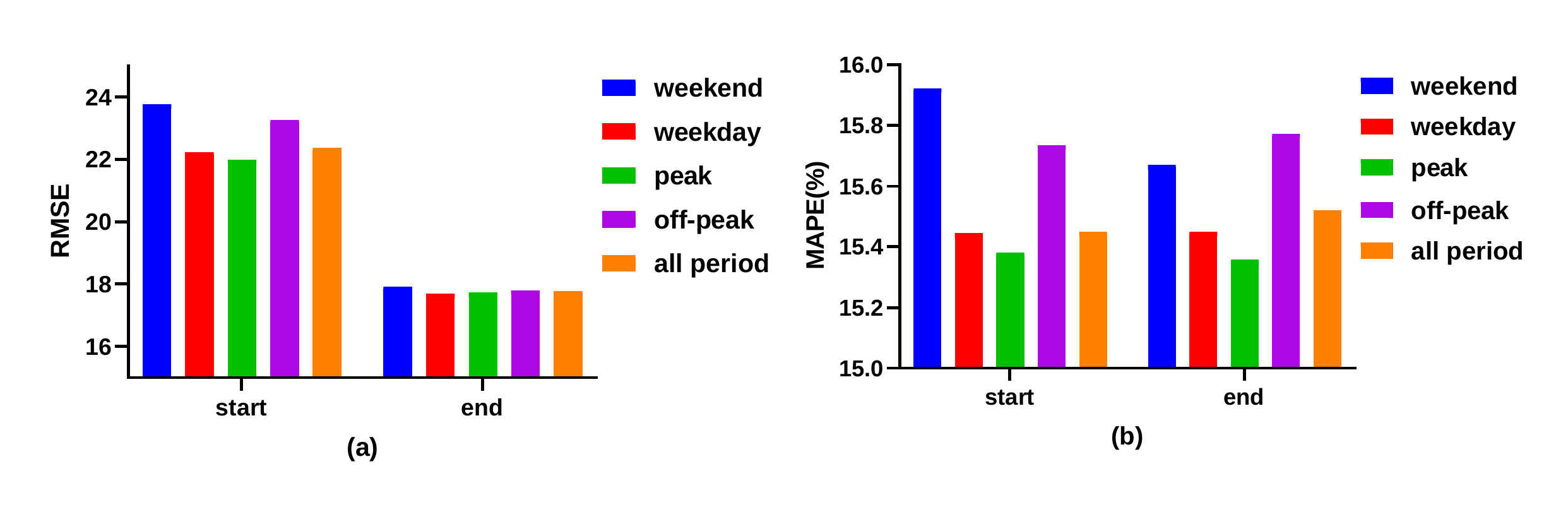} 
\caption{The experimental results of our integrated framework for different prediction periods on the NYC-Taxi dataset.}
\label{taxi-result-all-dif-period}
\end{figure}

We deduce the main reason that bicycle and taxi usage rates are mainly affected by weather and other external characteristics. For example, when there is heavy rainfall or high UV level, the bicycle usage rate will be reduced significantly. Besides, in extreme weather conditions, the possibility of taking a taxi will dramatically increase. At this time, the influence of weather features on the prediction results of traffic data is undeniable. Moreover, the trends of traffic volume on holidays, weekends, and weekdays are much different. Thus these factors affect the accuracy of prediction significantly. Figures \ref{bike-result-all-dif-period} and \ref{taxi-result-all-dif-period} show our framework for predicting RMSE and MAPE of traffic volume in different periods on the NYC-Bike and NYC-Taxi dataset, respectively. So we believe that the framework can obtain the superior performance on predicting the traffic conditions of weekdays.



\textbf{BDLSTM:} From Tables \ref{tab:bike} and \ref{tab:taxi}, we can see that BDLSTM has a better performance than LSTM. The reason is that in the chain-like gated structure of BDLSTM, the forward and backward dependencies of traffic volume are taken into consideration. In other words, BDLSTM can provide double protection, preventing critical information from being filtered out and ensuring them to pass through the chain-like gated structures efficiently.
Besides, due to the periodicity of traffic data, BDLSTM can learn the forward feature dependencies in chronological order and the reverse feature dependencies in reverse chronological order.
\begin{figure}[!ht]
\centering
\includegraphics[width=0.95\columnwidth]{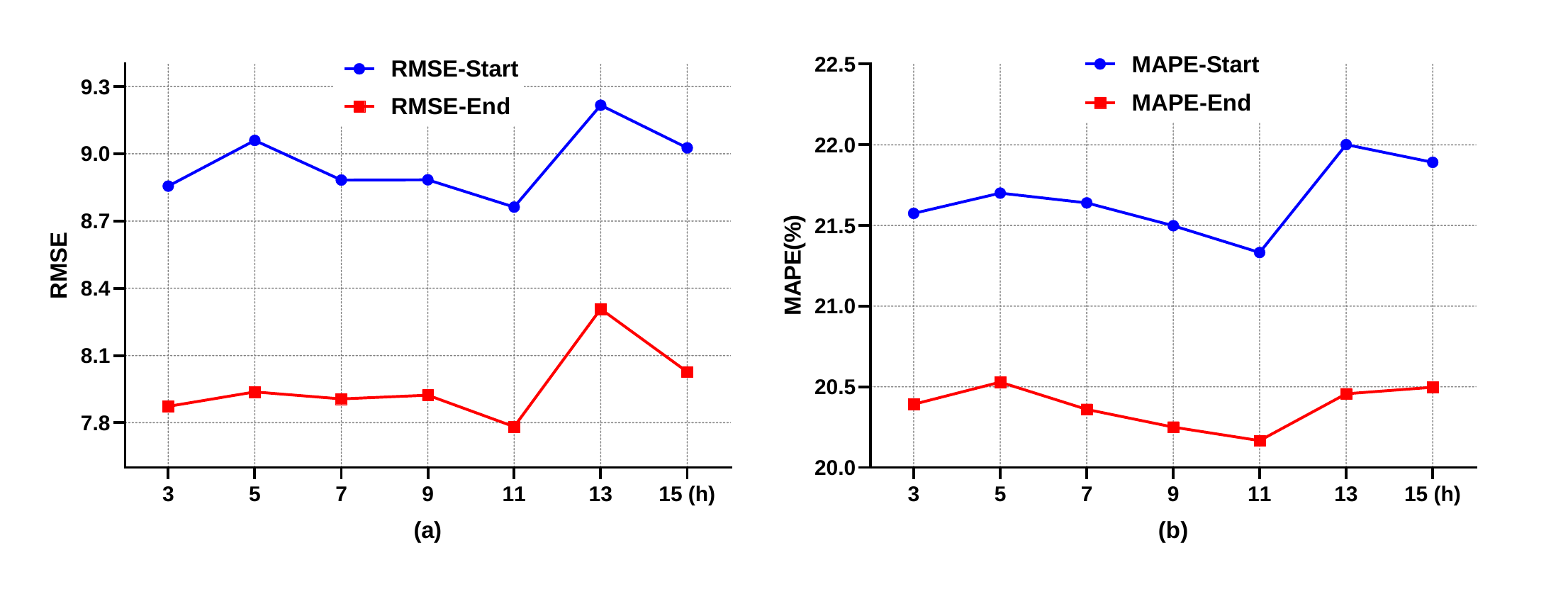} 
\caption{The RMSE and MAPE of the LSTM with attention mechanism on the NYC-Bike dataset.}
\label{bike-peak}
\end{figure}
\begin{figure}[!ht]
\centering
\includegraphics[width=0.95\columnwidth]{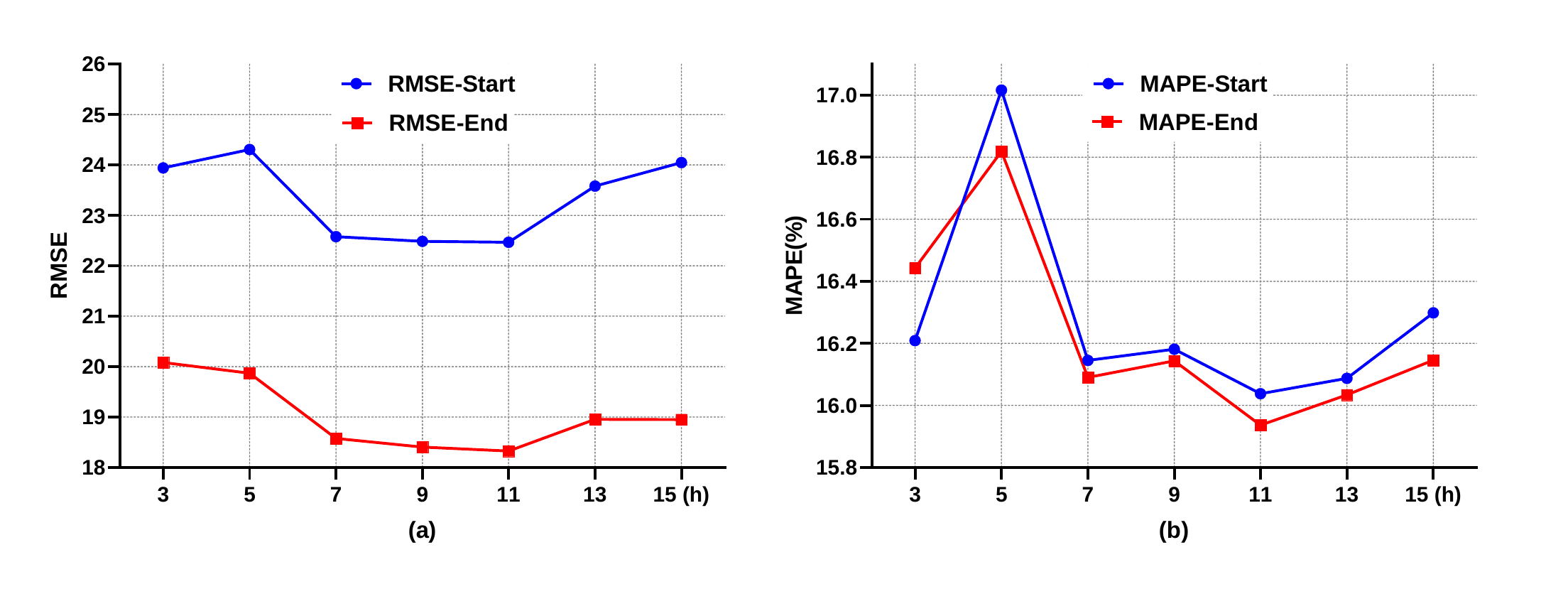} 
\caption{The RMSE and MAPE of the LSTM with attention mechanism on the NYC-Taxi dataset.}
\label{taxi-peak}
\end{figure}

\textbf{New Attention Mechanism:} 
It can be seen from the experimental results that LSTM with the new attention mechanism has a better performance than STDN \cite{Yao:2019tz}. The error rates at the start and end of the traffic volume have been reduced after the new attention mechanism is adopted. One reason is that we have added more information to the attention mechanism, including hourly (e.g., $\tau$ intervals), daily, and weekly periodic dependence. The other reason is that we have considered the relationship between the morning peak and the evening peak of the daily traffic volume. We do a lot of data analysis and experiments on the selection of peak intervals, and the experimental results are shown in Figures \ref{bike-peak} and \ref{taxi-peak}. When the value of the peak interval is set to 11 hours, the values of RMSE and MAPE at the start and the end of the period are minimal. It concludes that the traffic volume at the target region at time $t$ is related to the traffic conditions of the corresponding peak time.\\

\subsection{Time Complexity}
The most time-consuming part of the model is the capture of the temporal features dependency.
Due to the STDN uses LSTM and traditional attention mechanism, and our model uses the more complex gate structure of BDLSTM and the multi-scale attention mechanism, the time required for each iteration of the model is 1.52 times higher than that of the method STDN during the training, and the executing time of the model for prediction is 1.58 times higher than that of STDN.
We implement the proposed model in Python 3.6. The model has been successfully executed and tested on Ubuntu 18.04 platform, running on a PC with Intel Core CPU i7-7820X@3.60 GHz, 80 GB RAM, and Nvidia graphics card TITAN Xp. The training process of the model takes about 6.5 hours, and the prediction of the traffic volume takes about 3.3 minutes.

\section{Conclusion}
Many effective models have been designed to predict traffic conditions, where the latest model STDN is a combination of LSTM and attention mechanism with time intervals of days. In this paper, after analyzing the multi-scale correlation of traffic volume, we first propose a multi-scale modeling scheme, which contains the four levels of time intervals, hours, days, and weeks to improve the accuracy of traffic volume forecast during peak hours. This multi-scale model not only is useful for traffic volume in un-peak periods, but also can make more accurate predictions for peak periods. Secondly, we propose a way to combine the multi-scale attention mechanism and the BDLSTM. Finally, the multi-source information is added as the external features to improve the prediction accuracy of traffic volume. The experimental results on the NYC-Bike and NYC-Taxi show that the model achieves a more accurate prediction performance than the method STDN. Moreover, since our model is superior to the five baselines in terms of all the evaluation metrics on the different traffic datasets, we believe that it has generalization ability to a certain extent and can be applied to other time series prediction tasks. In the future, we plan to improve the real-time performance of the model so that it can be used to predict the actual road conditions.

\begin{acknowledgements}
This research is supported by the National Natural Science Foundation of China (NSFC 61572005, 61672086, 61702030, 61771058).
\end{acknowledgements}

%
%

\bibliographystyle{spmpsci}      
\bibliography{reference}   

\begin{thebibliography}{10}
\providecommand{\url}[1]{{#1}}
\providecommand{\urlprefix}{URL }
\expandafter\ifx\csname urlstyle\endcsname\relax
  \providecommand{\doi}[1]{DOI~\discretionary{}{}{}#1}\else
  \providecommand{\doi}{DOI~\discretionary{}{}{}\begingroup
  \urlstyle{rm}\Url}\fi

\bibitem{BotoGiralda:2010jd}
Boto-Giralda, D., D{\'\i}az-Pernas, F.J., Gonz{\'a}lez-Ortega, D.,
  D{\'\i}ez-Higuera, J.F., Ant{\'o}n-Rodr{\'\i}guez, M.,
  Mart{\'\i}nez-Zarzuela, M., Torre-D{\'\i}ez, I.: {Wavelet-based denoising for
  traffic volume time series forecasting with self-organizing neural networks}.
\newblock Computer-Aided Civil and Infrastructure Engineering \textbf{25}(7),
  530--545 (2010)

\bibitem{Cui:2018traffic}
Cui, Z., Henrickson, K., Ke, R., Wang, Y.: {Traffic graph convolutional
  recurrent neural network: a deep learning framework for network-scale traffic
  learning and forecasting}.
\newblock CoRR \textbf{abs/1802.07007}, 1--11 (2018)

\bibitem{Cui:2018deep}
Cui, Z., Ke, R., Wang, Y.: {Deep bidirectional and unidirectional LSTM
  recurrent neural network for network-wide traffic speed prediction}.
\newblock CoRR \textbf{abs/1801.02143}, 1--12 (2018)

\bibitem{Dougherty:1997vt}
Dougherty, M.S., Cobbett, M.R.: {Short-term inter-urban traffic forecasts using
  neural networks}.
\newblock International Journal of Forecasting \textbf{13}(1), 21--31 (1997)

\bibitem{Greenberg:1959hg}
Greenberg, H.: {An analysis of traffic flow}.
\newblock Operations Research \textbf{7}(1), 79--85 (1959)

\bibitem{Guo:2010fb}
Guo, J., Williams, B.M.: {Real-time short-term traffic speed level forecasting
  and uncertainty quantification using layered kalman filters}.
\newblock Transportation Research Record \textbf{2175}(1), 28--37 (2010)

\bibitem{Guo2019AttentionBS}
Guo, S., Lin, Y., Feng, N., Song, C., Wan, H.: {Attention based
  spatial-temporal graph convolutional networks for traffic flow forecasting}.
\newblock Proceedings of the AAAI Conference on Artificial Intelligence
  \textbf{33}, 922--929 (2019)

\bibitem{Jeong:ww}
Jeong, Y.S., Byon, Y.J., Castro-Neto, M.M., Easa, S.M.: {Supervised
  weighting-online learning algorithm for short-term traffic flow prediction}.
\newblock IEEE Transactions on Intelligent Transportation Systems
  \textbf{14}(4), 1700--1707 (2013)

\bibitem{LeCun:2015wc}
LeCun, Y., Bengio, Y., Hinton, G.E.: {Deep learning}.
\newblock Nature \textbf{521}(7553), 436--444 (2015)

\bibitem{Luo:2019hm}
Luo, X., Li, D., Yang, Y., Zhang, S.: {Spatiotemporal traffic flow prediction
  with KNN and LSTM}.
\newblock Journal of Advanced Transportation \textbf{2019}, 10 (2019)

\bibitem{Lv:2015dl}
Lv, Y., Duan, Y., Kang, W., Li, Z., Wang, F.: {Traffic flow prediction with big
  data: a deep learning approach}.
\newblock IEEE Transactions on Intelligent Transportation Systems
  \textbf{16}(2), 865--873 (2015)

\bibitem{Shi:2015uua}
Shi, X., Chen, Z., Wang, H., Yeung, D.Y., Wong, W.K., Woo, W.c.: {Convolutional
  LSTM network: a machine learning approach for precipitation nowcasting}.
\newblock In: Advances in Neural Information Processing Systems, pp. 802--810
  (2015)

\bibitem{Silver:2016tu}
Silver, D., Huang, A., Maddison, C.J., Guez, A., Sifre, L., van~den Driessche,
  G., Schrittwieser, J., Antonoglou, I., Panneershelvam, V., Lanctot, M.,
  Dieleman, S., Grewe, D., Nham, J., Kalchbrenner, N., Sutskever, I.,
  Lillicrap, T., Leach, M., Kavukcuoglu, K., Graepel, T., Hassabis, D.:
  {Mastering the game of go with deep neural networks and tree search}.
\newblock Nature \textbf{529}, 484--489 (2016)

\bibitem{Sun:2015wj}
Sun, Y., Leng, B., Guan, W.: {A novel wavelet-SVM short-time passenger flow
  prediction in Beijing subway system}.
\newblock Neurocomputing \textbf{166}, 109--121 (2015)

\bibitem{Szeto:2009bi}
Szeto, W.Y., Ghosh, B., Basu, B., O{\textquoteright}Mahony, M.: {Multivariate
  traffic forecasting technique using cell transmission model and SARIMA
  model}.
\newblock Journal of Transportation Engineering \textbf{135}(9), 658--667
  (2009)

\bibitem{Tong:2017vy}
Tong, Y., Chen, Y., Zhou, Z., Chen, L., Wang, J., Yang, Q., Ye, J., Lv, W.:
  {The simpler the better: a unified approach to predicting original taxi
  demands based on large-scale online platforms}.
\newblock In: Proceedings of the 23rd ACM SIGKDD International Conference on
  Knowledge Discovery and Data Mining, pp. 1653--1662. ACM (2017)

\bibitem{van2012short}
Van~Lint, J., Van~Hinsbergen, C.: {Short-term traffic and travel time
  prediction models}.
\newblock Artificial Intelligence Applications to Critical Transportation
  Issues \textbf{22}(1), 22--41 (2012)

\bibitem{Wang:2017bt}
Wang, D., Cao, W., Li, J., Ye, J.: {DeepSD: supply-demand prediction for online
  car-hailing services using deep neural networks}.
\newblock In: 2017 IEEE 33rd International Conference on Data Engineering
  (ICDE), pp. 243--254. IEEE (2017)

\bibitem{Williams:2003br}
Williams, B.M., Hoel, L.A.: {Modeling and forecasting vehicular traffic flow as
  a seasonal ARIMA process: theoretical basis and empirical results}.
\newblock Journal of Transportation Engineering \textbf{129}(6), 664--672
  (2003)

\bibitem{Wu:2016vp}
Wu, Y., Tan, H.: {Short-term traffic flow forecasting with spatial-temporal
  correlation in a hybrid deep learning framework}.
\newblock CoRR \textbf{abs/1612.01022}, 1--14 (2016)

\bibitem{Xie:2007je}
Xie, Y., Zhang, Y., Ye, Z.: {Short-term traffic volume forecasting using kalman
  filter with discrete wavelet decomposition}.
\newblock Computer-Aided Civil and Infrastructure Engineering \textbf{22}(5),
  326--334 (2007)

\bibitem{Yao:2019tz}
Yao, H., Tang, X., Wei, H., Zheng, G., Li, Z.: {Revisiting spatial-temporal
  similarity: a deep learning framework for traffic prediction}.
\newblock Proceedings of the AAAI Conference on Artificial Intelligence
  \textbf{33}, 5668--5675 (2019)

\bibitem{Yao:2018wy}
Yao, H., Wu, F., Ke, J., Tang, X., Jia, Y., Lu, S., Gong, P., Ye, J., Li, Z.:
  {Deep multi-view spatial-temporal network for taxi demand prediction}.
\newblock In: Proceedings of the Thirty-Second AAAI Conference on Artificial
  Intelligence, (AAAI-18), the 30th innovative Applications of Artificial
  Intelligence (IAAI-18), and the 8th AAAI Symposium on Educational Advances in
  Artificial Intelligence (EAAI-18), New Orleans, Louisiana, USA, February 2-7,
  2018, pp. 2588--2595 (2018)

\bibitem{Yasdi:1999wt}
Yasdi, R.: {Prediction of road traffic using a neural network approach}.
\newblock Neural Computing {\&} Applications \textbf{8}(2), 135--142 (1999)

\bibitem{Ye:2019cp}
Ye, J., Sun, L., Du, B., Fu, Y., Tong, X., Xiong, H.: {Co-prediction of
  multiple transportation demands based on deep spatio-temporal neural
  network}.
\newblock In: Proceedings of the 25th ACM SIGKDD International Conference on
  Knowledge Discovery {\&} Data Mining, pp. 305--313. ACM (2019)

\bibitem{Yu:2018gn}
Yu, B., Yin, H., Zhu, Z.: {Spatio-temporal graph convolutional networks: a deep
  learning framework for traffic forecasting}.
\newblock In: Proceedings of the Twenty-Seventh International Joint Conference
  on Artificial Intelligence, \{IJCAI-18\}, pp. 3634--3640. IJCAI (2018)

\bibitem{Yu:2017tv}
Yu, H., Wu, Z., Wang, S., Wang, Y., Ma, X.: {Spatiotemporal recurrent
  convolutional networks for traffic prediction in transportation networks}.
\newblock Sensors \textbf{17}(7), 1501 (2017)

\bibitem{Zhang:2000dx}
Zhang, H.M.: {Recursive prediction of traffic conditions with neural network
  models}.
\newblock Journal of Transportation Engineering \textbf{126}(6), 472--481
  (2000)

\bibitem{Zhang:2017tq}
Zhang, J., Zheng, Y., Qi, D.: {Deep spatio-temporal residual networks for
  citywide crowd flows prediction}.
\newblock In: Proceedings of the Thirty-First AAAI Conference on Artificial
  Intelligence, pp. 1655--1661 (2017)

\bibitem{Zhang:2018uj}
Zhang, J., Zheng, Y., Qi, D., Li, R., Yi, X., Li, T.: {Predicting citywide
  crowd flows using deep spatio-temporal residual networks}.
\newblock Artificial Intelligence \textbf{259}, 147--166 (2018)

\bibitem{Zhao:2018vv}
Zhao, L., Song, Y., Deng, M., Li, H.: {Temporal graph convolutional network for
  urban traffic flow prediction method}.
\newblock CoRR \textbf{abs/1811.05320}, 1--10 (2018)

\end{thebibliography}


\end{document}